\pdfoutput=1
\documentclass[11pt]{article}
\usepackage{booktabs}
\usepackage{enumitem}
\usepackage{ragged2e}
\usepackage{array}

\newcolumntype{P}[1]{>{\RaggedRight\arraybackslash}p{#1}}
\usepackage{dblfloatfix}

\usepackage[final]{acl}

\usepackage{times}
\usepackage{latexsym}

\usepackage[T1]{fontenc}
\usepackage{amsfonts}

\usepackage[utf8]{inputenc}

\usepackage{microtype}

\usepackage{inconsolata}
 \usepackage{amsmath} 
\usepackage{graphicx}

\usepackage{float}

\title{CausalRAG: Integrating Causal Graphs into Retrieval-Augmented Generation}

\author{
 \textbf{Nengbo Wang\textsuperscript{1,2}},
 \textbf{Xiaotian Han\textsuperscript{1}},
 \textbf{Jagdip Singh\textsuperscript{2}},
 \textbf{Jing Ma\textsuperscript{1}},
 \textbf{Vipin Chaudhary\textsuperscript{1}}
\\
 \textsuperscript{1}Department of Computer and Data Sciences, Case Western Reserve University\\
 \textsuperscript{2}Department of Design and Innovation, Case Western Reserve University
\\
 \href{mailto:nengbo.wang@case.edu}{nengbo.wang@case.edu}
}

\begin{document}
\maketitle
\begin{abstract}
Large language models (LLMs) have revolutionized natural language processing (NLP), particularly through Retrieval-Augmented Generation (RAG), which enhances LLM capabilities by integrating external knowledge. However, traditional RAG systems face critical limitations, including disrupted contextual integrity due to text chunking, and over-reliance on semantic similarity for retrieval. To address these issues, we propose \textit{CausalRAG}, a novel framework that incorporates causal graphs into the retrieval process. By constructing and tracing causal relationships, \textit{CausalRAG} preserves contextual continuity and improves retrieval precision, leading to more accurate and interpretable responses. We evaluate \textit{CausalRAG} against regular RAG and graph-based RAG approaches, demonstrating its superiority across several metrics. Our findings suggest that grounding retrieval in causal reasoning provides a promising approach to knowledge-intensive tasks\footnote{Code is publicly available at \url{https://github.com/Pwnb/CausalRAG}.}. 

\end{abstract}

\section{Introduction}

The rapid advancements in large language models (LLMs) have revolutionized the field of natural language processing (NLP), enabling a wide range of applications \cite{anthropic_claude_2024,google_our_2024,openai_hello_2024}. However, their reliance on pre-trained knowledge limits their ability to integrate and reason over dynamically updated external information, particularly in knowledge intensive domains such as academic research. Retrieval-Augmented Generation (RAG) has emerged as a promising framework to address this limitation \cite{lewis_retrieval-augmented_2021}, combining retrieval mechanisms with generative capabilities to enhance contextual understanding and response quality.  

Recent research has focused on improving RAG along two primary directions: 1) enhancing retrieval efficiency and integration mechanisms by designing more adaptive and dynamic retrieval frameworks \cite{gan2024similarityneedendowingretrieval, ravuru2024agenticretrievalaugmentedgenerationtime, zhang2024raftadaptinglanguagemodel}; 2) improving the representation of external knowledge to facilitate retrieval and reasoning, with graph-based RAGs being a dominant approach \cite{edge_local_2024, guo_lightrag_2024, potts_lazygraphrag_2024}. Despite these advancements, existing RAG architectures still face critical limitations that impact retrieval quality and response accuracy, primarily due to three key issues: 1) disruption of contextual integrity caused by the text chunking design; 2) reliance on semantic similarity rather than causal relevance for retrieval; and 3) a lack of accuracy in selecting truly relevant documents. 

Through a combination of theoretical analysis and empirical evaluation, we rethink the limitations of current RAG systems by introducing a novel perspective based on context recall and precision metrics. Our findings reveal that both regular and graph-based RAGs struggle not only to retrieve truly grounded context but also to accurately discern the relationship between retrieved content and the user query. We identify this fundamental issue as one primary reason why LLMs in RAG frameworks often generate \textbf{seemingly relevant yet shallow responses that lack essential details}.

To address these gaps, we introduce \textit{CausalRAG}, a novel RAG framework that integrates causal graphs to enhance retrieval accuracy and reasoning performance. Unlike regular and graph-based RAGs, \textit{CausalRAG} explicitly identifies causal relationships within external knowledge, preserving contextual coherence while capturing underlying cause–effect dependencies. By ensuring that retrieved documents are both relevant and causally grounded, \textit{CausalRAG} \textbf{enables the generation of more contextually rich and causally detailed responses}. This approach not only improves retrieval effectiveness but also mitigates hallucinations and enhances answer faithfulness.  

We evaluate \textit{CausalRAG} on datasets from diverse domains and across varying context lengths, comparing its performance with regular RAG and other graph-based RAG frameworks. Our experiments assess performance across three key metrics: answer faithfulness, context recall, and context precision. Results demonstrate that \textit{CausalRAG} achieves superior performance across different contexts. Additionally, we conduct a case study and a parameter analysis to further examine our framework, analyzing and providing insights that contribute to ongoing research in RAG. The contributions of this work are threefold:  
\begin{itemize}
    \item We systematically identify the inherent limitations of RAG’s retrieval process through analytical and experimental study. We uncovered the major reason why LLMs in RAG tend to generate superficial, generalized answers that lack the grounded details expected by users.
    \item We propose \textit{CausalRAG}, a framework that enhances both retrieval and generation quality by incorporating causality into the RAG, effectively addressing these limitations.
    \item Our work further mitigates hallucination issues and significantly improves the interpretability of AI systems. We summarize key findings and insights in both retrieval and generation process, contributing to RAG research.
\end{itemize}

\section{Related Work}

\subsection{Retrieval-Augmented Generation}
RAG enhances LLMs’ ability to handle knowledge-intensive tasks by integrating external knowledge retrieval \cite{lewis_retrieval-augmented_2021}. Existing research primarily advances RAG along two key dimensions: 1) improving retrieval efficiency and integration mechanisms; 2) enhancing the representation and multi-stage utilization of external knowledge—particularly through knowledge graphs—to support deeper reasoning and generation.

\noindent\textbf{Optimizing Retrieval Flow and Interaction.} The first stream focuses on improving the interaction flow within the RAG system to enhance output quality. Approaches have introduced pre-retrieval, retrieval, and post-retrieval refinements to mitigate redundancy and computational overhead \cite{wang_speculative_2024}. Modular RAG architectures further advance this by enabling iterative retrieval-generation cycles, allowing dynamic interactions between retrieval and content creation. For example, CAiRE-COVID \cite{su-etal-2020-caire} demonstrated the effectiveness of iterative retrieval in multi-document summarization, while some work \cite{feng2023retrievalgenerationsynergyaugmentedlarge} extended this approach to multi-hop question answering. Recent innovations include METRAG \cite{gan2024similarityneedendowingretrieval}, which integrates LLM supervision to generate utility-driven retrieval processes, and RAFT \cite{zhang2024raftadaptinglanguagemodel}, which trains models to disregard distractor documents while improving citation accuracy through chain-of-thought reasoning.

\noindent\textbf{Structuring and Utilizing External Knowledge for Multi-stage Reasoning.}
The second stream focuses on representing and processing external knowledge—often in the form of knowledge graphs—to improve retrieval depth and reasoning accuracy. GraphRAG \cite{edge_local_2024} treats documents as interconnected nodes to capture thematic and causal links, while LightRAG \cite{guo_lightrag_2024} adopts a lightweight, dual-level retrieval approach for dynamic updates. Lazy GraphRAG \cite{potts_lazygraphrag_2024} further improves efficiency by delaying heavy computations until query time.

More recently, several multi-stage RAG frameworks have emerged that explicitly structure the retrieval and generation process across multiple levels. PolyRAG \cite{chen_knowledge_2025} builds a hierarchical knowledge pyramid, allowing coarse-to-fine retrieval over different abstraction layers. KG\textsuperscript{2}RAG \cite{zhu_knowledge_2025} guides RAG with KG-guided chunk expansion process to enhance factual grounding and control. GenTKGQA \cite{gao_two-stage_2024} employs a two-stage pipeline to first retrieve temporal subgraphs and then generate based on virtual knowledge indicators. Among them, HippoRAG2 \cite{gutierrez_rag_2025} focuses on combining PageRank algorithm with deeper passage integration, improving both knowledge retention and answer quality. However, most of these approaches focus primarily on \textbf{matching and ranking information within the knowledge graph, with limited attention to the causal intent behind the user's query.} A key challenge in RAG remains ensuring that retrieved information is not only relevant but also coherently aligned with the user’s underlying reasoning needs and the generation objective \cite{gupta_comprehensive_2024}.

\begin{figure*}[h]
  \includegraphics[width=1\linewidth, trim=0mm 0mm 0mm 0mm, clip]{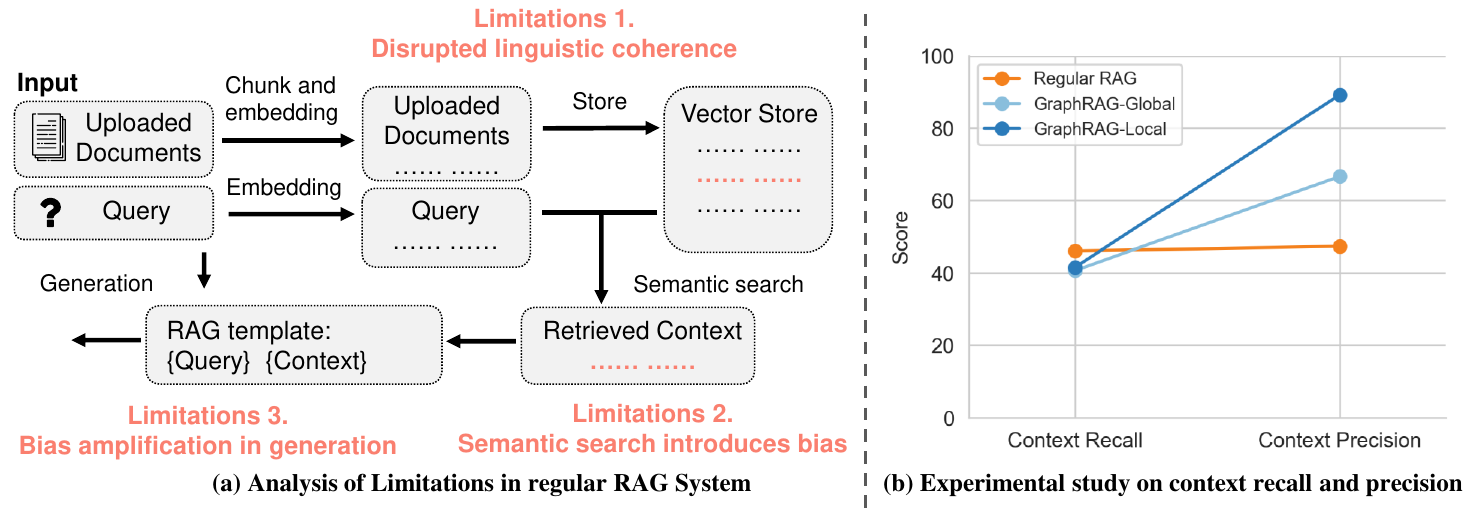} \hfill
  \captionsetup{aboveskip=0pt, belowskip=5pt}
  \caption {Analytical and experimental studies reveal limitations in regular RAG and GraphRAG. (a) identifies three key retrieval and generation issues in regular RAG; (b) evaluates RAG via context precision and recall, showing regular RAG excels in recall but lacks precision. GraphRAG improves precision but trades off some recall.}
  \label{fig1:workflow}
\end{figure*}

\subsection{Causal Graphs and RAG}

The combination of causal graphs and RAG has emerged as a promising approach to enhancing knowledge retrieval and reasoning. As causality provides a structural understanding of dependencies within data, it enables more interpretable and reliable AI outputs \cite{ma_causal_2024}. Existing research in this domain primarily advances causal discovery with RAG and LLMs. For instance, some work proposed an LLM-assisted breadth-first search (BFS) method for full causal graph discovery, significantly reducing time complexity \cite{jiralerspong_efficient_2024}. Additionally, some further introduced a correlation-to-causation inference (Corr2Cause) task to evaluate LLMs' ability to infer causation from correlation, revealing their limitations in generalization across different datasets \citep{jin_can_2024}.

Despite these advancements, most studies focus on utilizing RAG or LLMs for causal discovery or causal effect estimation \cite{ma_causal_2024, kiciman_causal_2024}, whereas \textbf{the direct integration of causal graphs into RAG architectures remains largely unexplored}. Our work aims to be a pioneer in this direction. A few existing studies have touched upon this concept but differ in scope. One approach integrates causal graphs within the LLM architecture itself, structuring the transformer’s internal token processing using causality rather than enhancing RAG retrieval \cite{chen_rethinking_2024}. Another employs causal graphs in RAG systems but focuses on the pre-retrieval stage and largely reduces the core process into a single embedding model without deeper exploration \cite{samarajeewa_causal_2024}. GraphRAG \cite{edge_local_2024}, while influential for its use of community detection and graph summarization, does not incorporate causality—nor do many recent multi-stage graph-based RAG frameworks.

In the following sections, we first analyze the nature of regular RAG and grah-based RAGs from a novel perspective, identifying their inherent limitations. We then introduce a causal graph structure to address these gaps.

\section{Why Regular RAG Fails in Providing Accurate Responses}

In this section, through both analytical and experimental investigations, we identify three fundamental limitations of regular RAG and rethink its design by examining its three core elements—user query, retrieved context, and response—through a novel perspective based on precision and recall.

\subsection{Limitations of Regular RAG}
The first limitation arises from RAG’s common practice of chunking texts into minimal units (as illustrated in Figure \ref{fig1:workflow}a). This process disrupts the natural linguistic and logical connections in the original text. These connections are crucial for maintaining contextual integrity, and if they are lost, an alternative mechanism must be implemented to restore them. 

The second limitation lies in the semantic search process. RAG typically retrieves the semantically closest documents from a vector database based on query similarity. However, in many cases, critical information necessary for answering a query is not semantically similar but rather causally relevant. A classic example is the relationship between diapers and beer—while they are not semantically related, they may exhibit a causal connection in real worlds. This limitation suggests that RAG’s reliance on semantic similarity may lead to the retrieval of contextually irrelevant but superficially related information.

The third limitation is that even when RAG retrieves a relevant context, this does not necessarily guarantee an accurate response. To formalize this issue, we used two key metrics: \textit{context recall} and \textit{context precision}, defined as follows:

\begin{equation}
\text{Context Recall} = \frac{\sum_{i=1}^{N} \mathbb{I} (C_i \in R)}{|R|}    
\end{equation}

Here $R$ is the reference set of all relevant references. $C_i$ is the $i^{th}$ retrieved reference. $\mathbb{I} (C_i \in R)$ is an indicator function that returns 1 if $C_i$ belongs to the reference set $R$, otherwise 0. It should always return 1 if no hallucination occurs in the LLM.

\begin{equation}
\text{Context Precision} = \frac{\sum_{i=1}^{N} \mathbb{I_Q} (C_i \in R)}{\sum_{i=1}^{N} \mathbb{I} (C_i \in R)}
\end{equation}

Here $\mathbb{I_Q} (C_i \in R)$ is an indicator function that returns 1 if the context retrieved is causally related to the user's query, otherwise 0.

\medskip\noindent\textbf{Recall–precision Perspective.}  
Context recall measures the extent to which relevant contextual information can be retrieved from external knowledge given a query. In practice, increasing the number of retrieved documents typically improves recall in RAG systems. However, this semantic search–based approach often sacrifices precision—the proportion of retrieved content that is truly correct and directly relevant to the user query. For example, when asking, “How does this article define AI?”, regular RAG tends to retrieve all cited definitions that are semantically similar, even though only the author’s own definition is actually pertinent. This illustrates a core limitation of relying on semantic similarity rather than causal relevance: the retrieval of content that appears related but is logically irrelevant. More critically, low retrieval precision introduces systematic bias, diluting the accuracy and reliability of the retrieved context.

In summary, while RAG can recall numerous answers from reference materials, the proportion of correct context remains low, ultimately reducing its precision. This recall-precision perspective provides a new lens to see the limitations of RAG frameworks.

\subsection{Rethinking Graph-based RAGs} Applying this perspective, we can better understand why Graph-based RAGs serve as improved variants of RAG. By summarizing and ranking the importance of graph information before retrieval, they largely enhance the quality of retrieved context, thereby improving context precision. However, it only partially addresses the identified limitations, as its summarization process does not entirely filter out irrelevant information. More importantly, its reliance on subgraph summarization for retrieval may adversely impact recall by omitting less graph-central but still causally relevant context. To further examine these trade-offs, we conducted an experimental study to empirically validate these analytical insights.

\medskip\noindent\textbf{Experimental Study.} Figure \ref{fig1:workflow}(b) reports experimental results for Regular RAG and both variants of GraphRAG. The graph‑based approaches markedly improve context precision, reflecting the additional reranking and subgraph‑summary steps in their retrieval pipelines. This gain comes with a modest reduction in context recall, indicating that some relevant passages are pruned during summarization. These scores were obtained with the Ragas evaluation framework \cite{es2023ragasautomatedevaluationretrieval}; more comprehensive experiments and implementation details are shown in Section \ref{section5experiment}.

By combining our recall–precision-based analytical insights with empirical findings, we highlight the inherent limitations of both standard RAG and its graph-based extensions. Specifically, we show that relying on semantic similarity and subgraph summarization—rather than causal relationships—introduces trade-offs that often result in superficial and less accurate generation. We also present a case study in Section~\ref{section_case_study}, which explicitly compares the retrieval processes across different RAG frameworks. In the following section, we introduce our proposed framework, \textit{CausalRAG}, developed to address these limitations through causally grounded retrieval.

\begin{figure*}[t]
  \includegraphics[width=1\linewidth, trim=0mm 25mm 0mm 0mm, clip]{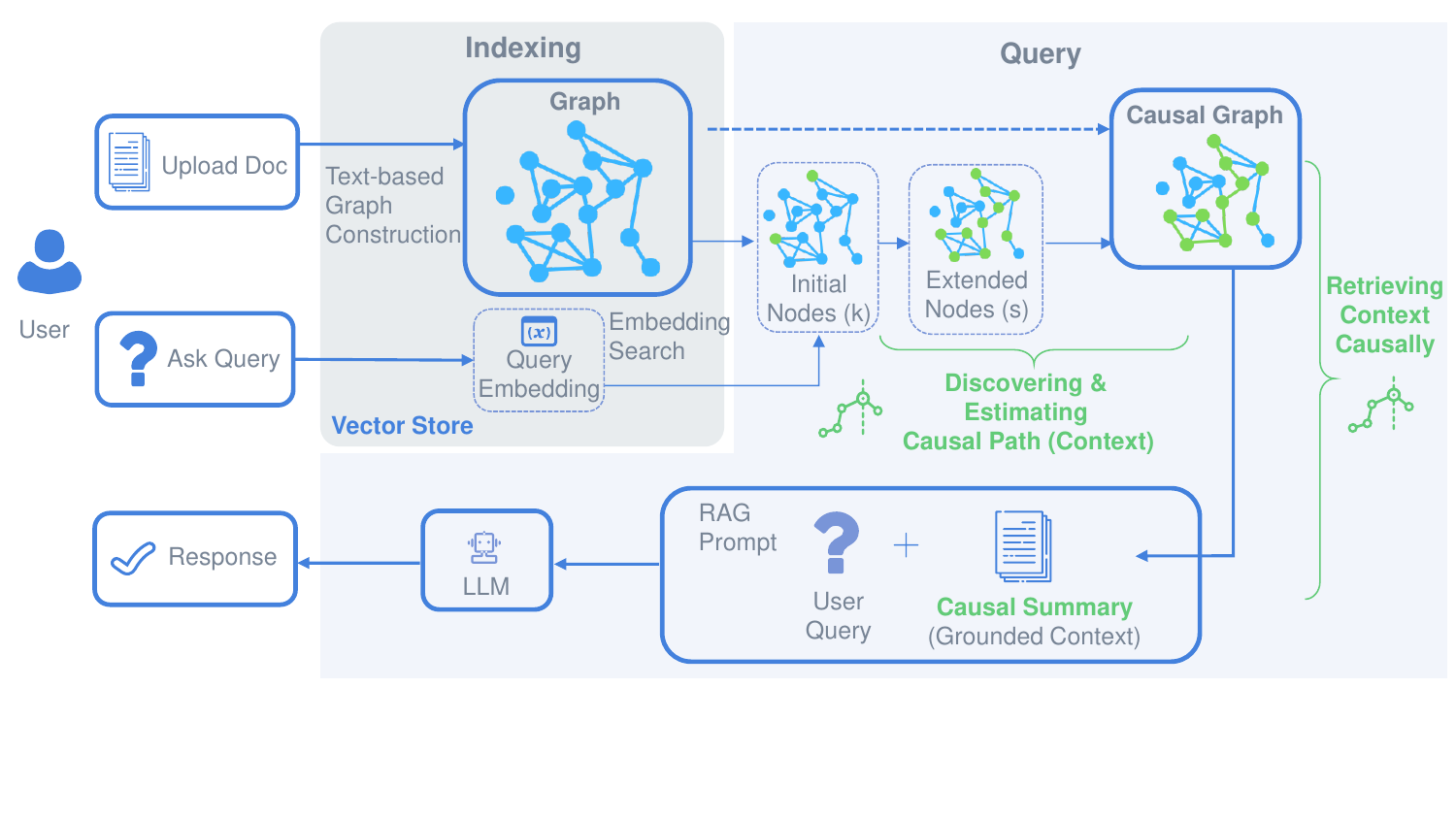} \hfill
  \captionsetup{aboveskip=0pt, belowskip=5pt}
  \caption {Overview of \textit{CausalRAG}'s architecture. Documents are indexed as graphs, and queries retrieve causally related nodes. A causal summary is generated and combined with the query to ensure grounded responses.}
  \label{fig2:causalragoverview}
\end{figure*}

\section{Methodology}
In this section, we introduce our proposed framework—\textit{CausalRAG}—which integrates RAG with causality to overcome the limitations of existing RAG systems. Overall, \textit{CausalRAG} constructs a text-based graph from uploaded documents and discovers causal paths among nodes to guide retrieval (Figure~\ref{fig2:causalragoverview}). By embedding the user query, matching relevant nodes, and expanding them along causally linked paths, the system generates a causal summary that serves as the retrieval context. This approach enables the framework to preserve contextual coherence and retrieve more targeted and causally relevant evidence. We now describe each step of the framework in detail.

\subsection{Indexing}  
At the outset, upon receiving the user's uploaded documents and query, the system first indexes both inputs into a vector database. For the documents, we employ a text-based graph construction method that transforms unstructured text into a structured graph comprising nodes and edges. Specifically, we follow the approach proposed in LangChain~\cite{chase_langchain_2022}, where an LLM parses the text to identify key entities or concepts as nodes and infers relationships between them as edges. Though this is a widely adopted approach in RAG research, we further validate the resulting graphs using expert knowledge, as discussed in the case study (Section~\ref{section_case_study}). Once the graph is constructed, it is embedded and stored in the vector database, enabling efficient similarity-based retrieval. The user query is also embedded at this stage, preparing the system for subsequent matching. Importantly, this indexing process is performed offline and independently of query time, ensuring fast and scalable inference.

\subsection{Discovering and Estimating Causal Paths}  
This is the first step during query time. We begin by matching the user query to nodes in the graph based on their embedding distance. The $k$ nodes with the smallest distances are selected, representing the most relevant information directly related to the query. Notably, $k$ is a tunable parameter—higher values retrieve more relevant information at the cost of increased computational complexity.  

After selecting the initial $k$ nodes, we expand the search along the base graph’s edges by a step size of $s$, thereby broadening the retrieved context. This step is crucial as it preserves causal and relational connections within the text, allowing \textit{CausalRAG} to retrieve more context while maintaining high recall. The parameter $s$ controls the depth of expansion, where higher values lead to more diverse information retrieval.  

Once the relevant nodes and edges are collected, we employ an LLM to identify and estimate causal paths within them, constructing a refined causal graph and generating the causal summary report (see LLM prompts in Appendix \ref{sec:appendixA}). LLMs have demonstrated superiority in discerning and analyzing causal relationships \cite{zhang_causal_2024,zhou_causalbench_2024} and this step ensures that \textit{CausalRAG} prioritizes causally relevant information, improving precision.

Furthermore, the derived causal graph serves two key purposes: 
1) It preserves causally relevant information that traditional retrieval methods struggle to capture. More importantly, by adjusting the parameter $s$, this approach can capture long-range causal relationships within the text, particularly when the text is lengthy;
2) It filters out semantically related but causally irrelevant information. Without this filtering, responses may contain unnecessary or even hallucinated content, compromising answer faithfulness.  

\begin{table*}
\centering
\begin{tabular}{lcccccc}
\hline
\textbf{Life Sciences} & \textbf{Computing \& Math} & \textbf{Social Sciences} & \textbf{Physics} & \textbf{Other} & \textbf{Total Tokens} \\
\hline
13.27\% & 14.29\% & 21.43\% & 14.29\% & 36.73\% & 21{,}285 \\
\hline
\end{tabular}
\caption{Statistics of the dataset domain distribution and token lengths.}
\label{tab:domain_token_distribution}
\end{table*}

\begin{figure*}[h]
    \centering
    \includegraphics[width=1\linewidth, trim=0mm 0mm 0mm 0mm, clip]{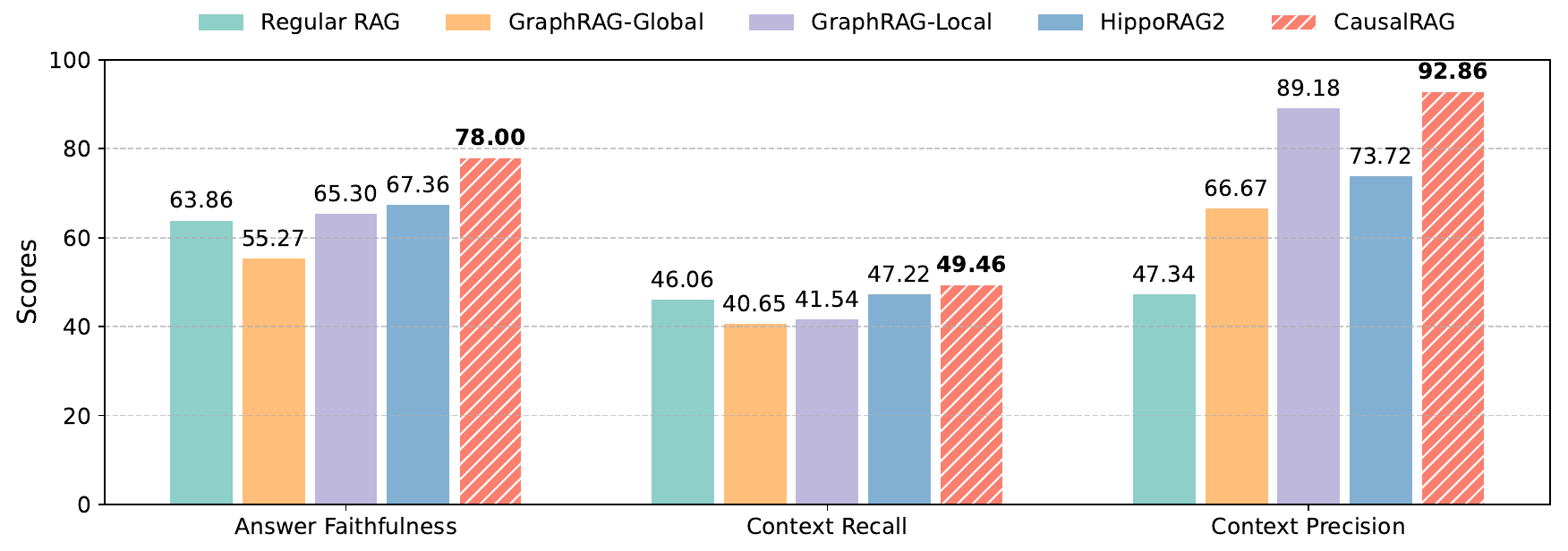} 
    \captionsetup{aboveskip=5pt, belowskip=5pt}
    \caption{Performance comparison of \textit{CausalRAG}, regular RAG, and other graph-based RAGs across three key metrics: answer faithfulness, context recall, and context precision.}
    \label{fig3:evaluation}
\end{figure*}

\subsection{Retrieving Context Causally}  
After constructing the causal graph, we summarize the retrieved information and generate a causal summary (see LLM prompts in Appendix~\ref{sec:appendixA}). Notably, the input at this stage consists of information that is not only highly relevant but also causally grounded in the user's query, ensuring greater validity. This approach contrasts with traditional retrieval methods, which often rely purely on semantic similarity and may retrieve contextually related yet causally irrelevant information.

The causal summary is derived by tracing key causal paths within the graph, prioritizing nodes and relationships that contribute directly to answering the query. This ensures that the retrieved information maintains logical coherence and factual consistency while filtering out spurious or weakly related context. Additionally, by leveraging causal dependencies, our method reduces the risk of retrieving semantically similar but misleading evidence.

Once the causal summary is generated, it is combined with the user query to construct a refined prompt for \textit{CausalRAG}. This structured final input allows RAG to focus on reasoning through causal relationships rather than merely aggregating loosely related text spans.

\section{Experiment} \label{section5experiment}

To evaluate the effectiveness of \textit{CausalRAG}, we conduct a series of experiments comparing it with regular RAG and other competitive graph-based RAGs across multiple performance metrics to ensure a comprehensive assessment. In addition, we present a case study that explicitly compares the retrieval processes of different RAG variants, and a parameter study to further examine the behavior and performance of \textit{CausalRAG}.

\subsection{Experimental Setup}  

\noindent\textbf{Baselines.} We evaluate five RAG variants: Regular RAG \cite{lewis_retrieval-augmented_2021}, GraphRAG with both local and global search \cite{edge_local_2024}, HippoRAG2 \cite{gutierrez_rag_2025}, and our proposed \textit{CausalRAG}. Regular RAG serves as a standard baseline, relying solely on semantic similarity for retrieval. GraphRAG is a widely recognized framework that leverages graph community summaries, and we include both of its modes: the local version retrieves from raw document graphs and is well-suited for passage-level queries, while the global version summarizes graph communities to support broader context understanding. HippoRAG2 is a recent and competitive method that enhances retrieval by selecting seed nodes and ranking filtered triples for generation.

\medskip\noindent\textbf{Datasets.}
While many open-domain QA benchmark datasets exist, most are designed for explicit fact retrieval (e.g., “When was Google founded?” “1998.”) and target classic NLP tasks. These datasets often fall short in evaluating discourse-level understanding, such as querying the underlying ideas, logic, or narrative within a document—tasks that better reflect real-world needs. Although some reading comprehension datasets exist \cite{rajpurkar_know_2018}, their answers are typically short entities and designed for NLP models rather than RAG systems.

To more effectively evaluate RAGs in knowledge-intensive tasks, recent research calls for datasets that require higher-level discourse understanding, such as those based on podcasts, news articles, or Wikipedia \cite{edge_local_2024, gutierrez_rag_2025}. Following this direction, we use academic papers sampled from the OpenAlex dataset \cite{priem2022openalexfullyopenindexscholarly}. Dataset statistics are provided in Table~\ref{tab:domain_token_distribution}. For each document, we use an LLM to generate $n = 5$ grounded questions, ensuring they are explicitly answerable (see Appendix~\ref{sec:appendixB} for examples).

\medskip\noindent\textbf{Metrics and Implementation Details.}
We use the Ragas evaluation framework \cite{es2023ragasautomatedevaluationretrieval} to assess all models on three metrics: answer faithfulness, context precision, and context recall. The definitions of context precision and context recall are provided in the previous section. Answer faithfulness measures factual consistency on a scale from 0 to 100, with higher scores indicating closer alignment with reference documents. We use GPT-4o-mini as the base LLM for all frameworks. We set the parameters $k = s = 3$ for \textit{CausalRAG}, and use the same $k$ value for GraphRAG’s community-based retrieval, Regular RAG’s document retrieval, and HippoRAG2’s triple-based retrieval to ensure a fair comparison.

\begin{figure*}[t]
  \includegraphics[width=1\linewidth, trim=0mm 0mm 0mm 1mm, clip]{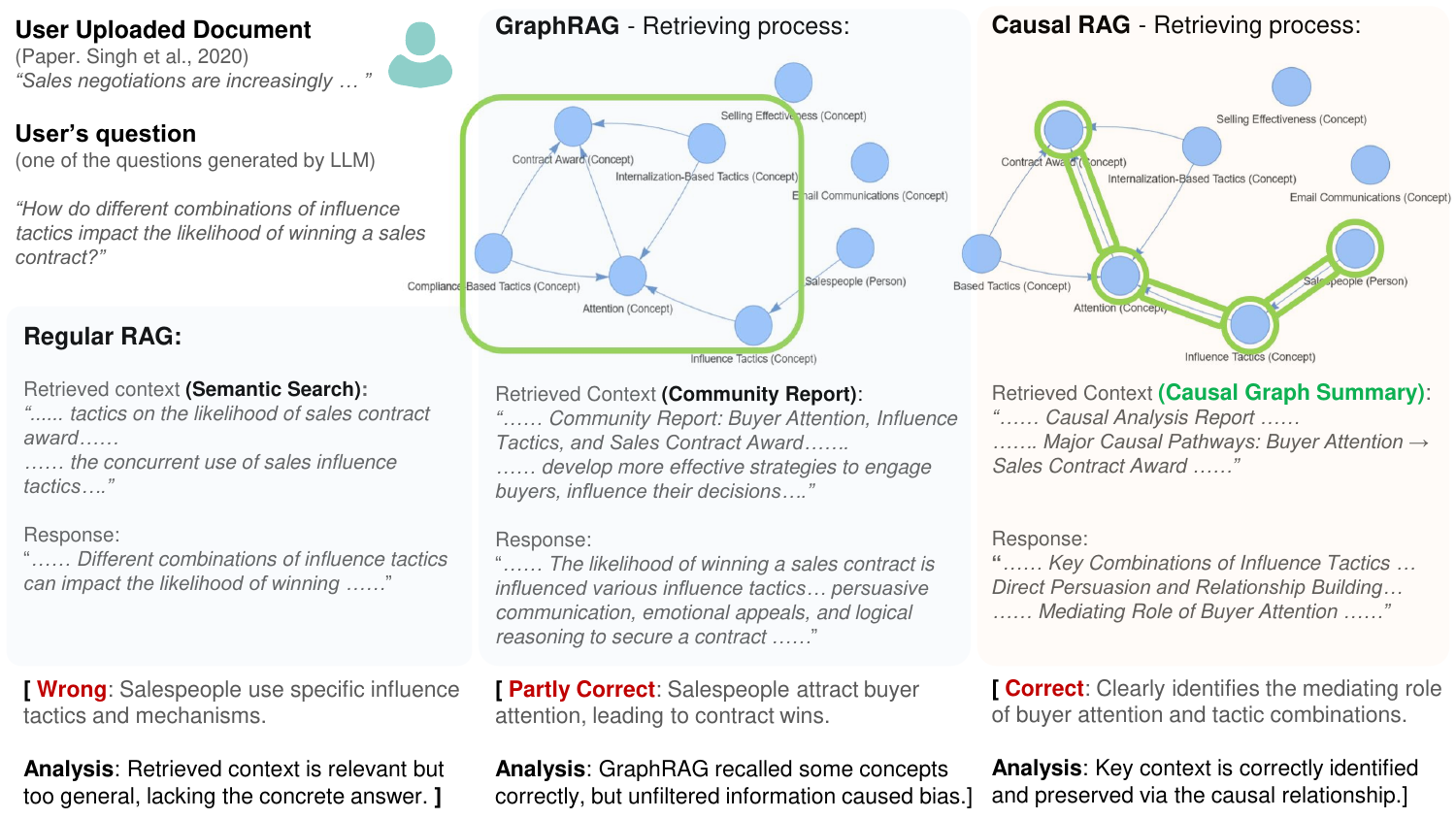} \hfill
  \captionsetup{aboveskip=0pt, belowskip=5pt}
  \caption {Case Study – A user uploads a long paper and asks a related question. This figure compares Regular RAG, GraphRAG, and \textit{CausalRAG} by analyzing their retrieval processes. It highlights the drawbacks of semantic and graph-based retrieval and shows how causal reasoning in CausalRAG leads to more robust and precise results.}
  \label{fig4:casestudyprocess}
\end{figure*}

\begin{figure}
    \centering
    \includegraphics[width=1\linewidth, trim=0mm 0mm 0mm 0mm, clip]{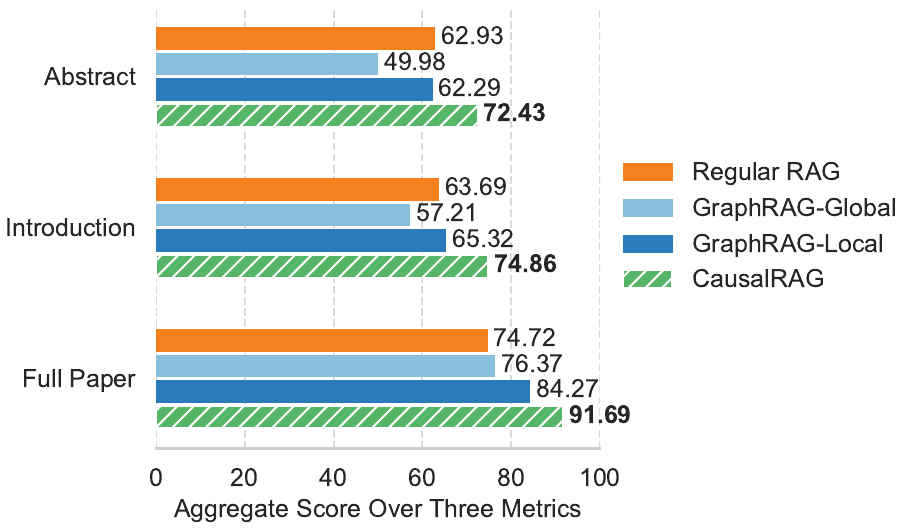}
    \captionsetup{aboveskip=5pt, belowskip=0pt}
    \caption{Case Study – A follow-up experiment evaluates the RAGs in the previous case using three versions of the same paper. Graph-based methods improve with length, while \textit{CausalRAG} remains consistently robust.}
    \label{fig5:casestudybarchart}
\end{figure}

\subsection{Performance Comparison}

Figure~\ref{fig3:evaluation} presents the main experimental results comparing five RAG frameworks across three evaluation metrics: answer faithfulness, context recall, and context precision. The results reveal distinct patterns in how different retrieval strategies affect generation quality. Below, we summarize the key findings and underlying trade-offs.

\medskip\noindent\textbf{Causality delivers the most balanced and accurate retrieval.}
The experimental results, as depicted in Figure \ref{fig3:evaluation}, reveal distinct strengths and weaknesses among the evaluated RAG frameworks, ultimately highlighting the advantages of integrating causal reasoning. \textit{CausalRAG} consistently achieves the most balanced and superior performance, leading in answer faithfulness ($78.00$), context precision ($92.86$), and maintaining competitive context recall ($49.46$). This suggests that by explicitly tracing and retrieving along causal paths, \textit{CausalRAG} effectively mitigates the common RAG pitfalls of relying purely on semantic similarity and grappling with contextual integrity.

\medskip\noindent\textbf{Graph‑level structuring improves precision but introduces a recall--coverage trade‑off.}
Graph-based RAG approaches, while an improvement over regular RAG in certain aspects, demonstrate a clear trade-off. GraphRAG-Local excels in context precision ($89.18$) but suffers from a low context recall ($41.54$). This indicates that while its direct querying of the raw knowledge graph yields highly relevant local information, it often discards broader supporting passages. The global variant, GraphRAG-Global, attempts to address this by summarizing community subgraphs, thereby improving recall ($47.22$). However, this comes at a significant cost to precision ($66.67$) and faithfulness ($55.27$). These results corroborate \cite{edge_local_2024}'s observation that finer‑grained community answers excel at factual grounding, whereas higher‑level summaries aid breadth but dilute specificity. In contrast, \textit{CausalRAG} manages to maintain high precision while achieving better recall through its targeted causal path selection.

\medskip\noindent\textbf{Entity‑centric multi‑stage KGs narrow the gap but still overlook discourse context.}
HippoRAG2 presents a strong performance, particularly in answer faithfulness ($67.36$) and context precision ($73.72$) when compared to Regular RAG and GraphRAG-Global, with a recall of $47.22$. Its methodology, which augments Personalized PageRank with LLM-filtered triples and passage nodes, clearly enhances retrieval quality. Nevertheless, its reliance on entity extraction leaves many context‑rich sentences unlinked to the query, a limitation they noted as concept--context trade‑off and need for deeper contextualization\cite{gutierrez_rag_2025}. Then we found \textit{CausalRAG} surpasses it across all metrics. By comparison, \textit{CausalRAG}'s causal path expansion allows for the integration of both entities and their underlying causal explanations. This approach yields a retrieval set that is not only broad in coverage but also more tightly aligned with the user's query, leading to more accurate and faithful responses by ensuring the retrieved context is causally pertinent rather than just semantically adjacent.

\subsection{Case Study}\label{section_case_study}

\noindent\textbf{Qualitative Exemplar.}
Figure \ref{fig4:casestudyprocess} traces the end‑to‑end behaviour of the three systems on a single question drawn from a marketing paper \cite{singh_business--business_2020}: “\emph{How do different combinations of influence tactics impact the likelihood of winning a sales contract?}”
The semantic baseline retrieves passages that mention “influence tactics” but stop short of establishing \emph{how} those tactics translate into contract wins; its answer therefore echoes the query without providing the missing mechanism.
Graph‑based retrieval narrows the search space by clustering entities such as \emph{buyer attention} and \emph{sales contract award}, yet the community summary also injects peripheral phrases, leading the model to an over‑generalised explanation that only partially matches the ground truth.
By contrast, the causality‑driven pipeline first aligns the query with a causal pathway—\emph{Influence Tactics} $\rightarrow$ \emph{Buyer Attention} $\rightarrow$ \emph{Contract Award}.
Because that pathway is preserved through retrieval and summarisation, the generated answer pinpoints the mediating role of buyer attention and specifies which tactic pairs are most effective, fully satisfying expert judgment.

\medskip \noindent\textbf{Length‑controlled Follow‑up.}
To test whether these qualitative differences persist when context grows, we replicate the experiment on three versions of the same paper—abstract (250 tokens), introduction (1k), and full text (16k)—and score each system on the composite of faithfulness, precision, and recall (Figure \ref{fig5:casestudybarchart}).
Graph‑level methods benefit from additional material: their aggregate score climbs from 49.98/62.29 on the abstract to 76.37/84.27 on the full paper, confirming that community‑based indexing scales gracefully with document length.
Yet the causality‑oriented approach remains consistently ahead, posting 72.43 on the abstract, 74.86 on the introduction, and 91.69 on the full text.
This steadiness indicates that causal expansion recovers the right evidence even when local term overlap is sparse (short documents) and still filters noise when semantic matches proliferate (long documents).

\medskip \noindent\textbf{Emerging Pattern.}
Taken together, the case study and its quantitative extension suggest a general rule: \emph{graph structure raises precision by adding relational cues, but explicit causal alignment is required to preserve both precision and recall across scales}.
Semantic search alone misses latent mechanisms; graph aggregation alone retains residual noise.
Embedding causal constraints into retrieval therefore offers a principled path to robust performance on knowledge‑intensive tasks, regardless of document length or query granularity.

\subsection{Parameter Study}

\begin{figure}
    \centering
    \includegraphics[width=1\linewidth, trim=0mm 0mm 0mm 0mm, clip]{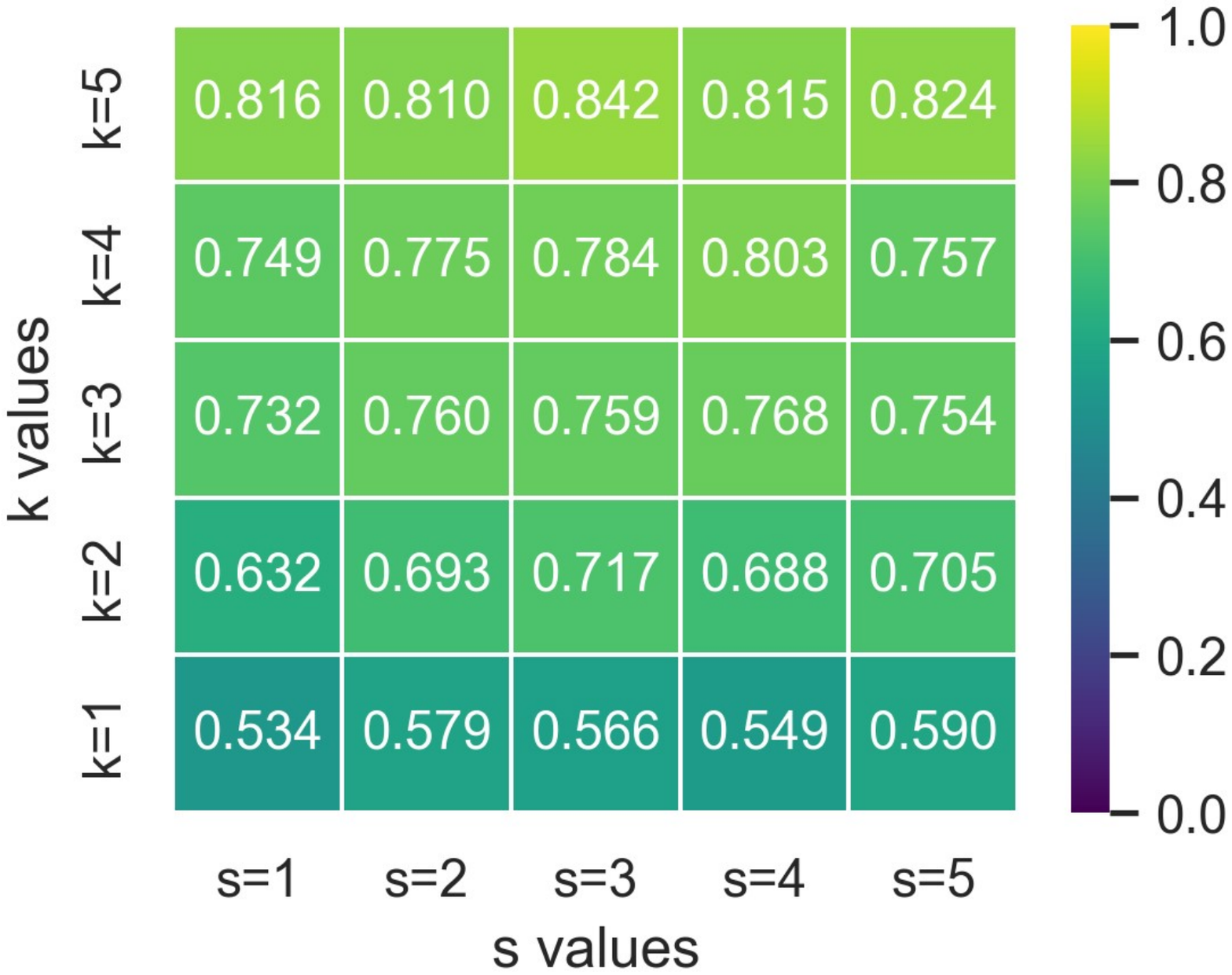}
    \captionsetup{aboveskip=5pt, belowskip=5pt}
    \caption{Parameter study showing how different parameter choices (k and s) affect model performance.}
    \label{fig6:parameterstudy}
\end{figure}

We also tested the impact of different parameter combinations of $k$ and $s$ on \textit{CausalRAG} (as shown in Figure \ref{fig6:parameterstudy}). Using the average of our evaluation metrics, we observe a consistent trend: the performance of \textit{CausalRAG} improves as $k$ and $s$ increase. Specifically, the performance rises from 0.534 at $k = s = 1$ to 0.824 at $k = s = 5$, aligning with intuitive expectations.

Notably, the improvement is more pronounced when increasing $k$ from 1 to 3, suggesting that retrieving additional context enhances reasoning quality. However, when $k \geq 4$, performance gains become less significant, indicating possible saturation due to information redundancy. Similarly, while increasing $s$ generally leads to better results, its effect diminishes at higher values of $k$, where retrieval is already extensive.

These results suggest an optimal trade-off between performance and computational efficiency. While the highest values ($k = 5, s = 5$) yield the best results, moderate settings such as $k = 3, s = 3$ still achieve competitive performance with lower retrieval costs. Future work could explore adaptive strategies to adjust these parameters dynamically based on query complexity.

\subsection{Conclusion and Future Work}

We introduced \textit{CausalRAG}, a novel framework that integrates causal reasoning into retrieval-augmented generation to address key limitations of existing RAG systems. Through theoretical analysis and empirical validation, we demonstrated that regular RAGs suffer from disrupted contextual integrity, semantic-over-causal retrieval, and context accuracy trade-offs. By leveraging causal graphs, \textit{CausalRAG} retrieves context that is not only relevant but causally grounded—enhancing generation quality, reducing hallucinations, and improving alignment between user queries and retrieved content. Our results show consistent improvements over strong baselines across diverse domains and context lengths.

Future work can extend \textit{CausalRAG} in several directions. An important future direction is to ideally evaluate scalability under long-context scenarios—requiring the construction and processing of numerous high-dimensional graphs, each derived from large-scale documents spanning millions of tokens. However, current practices often merge documents into a single graph or build smaller graphs for expert-verifiable segments, limiting real-world scalability. In addition, as LLMs evolve to support larger input sizes, developing scalable, graph-based causal retrieval methods for long-context reasoning should be an important next step.


\newpage
\section*{Limitations}
While \textit{CausalRAG} improves retrieval effectiveness through causal reasoning, it has certain limitations. First, the approach relies on the internal knowledge of LLMs to construct graphs and identify causal relationships. Although both capabilities have been actively studied, emerging domain-specific knowledge—such as in medicine or law—may still constrain its effectiveness in specialized contexts. Second, identifying causal paths during inference also requires additional LLM calls, introducing extra computational costs that may limit the performance in real-world deployments.

\section*{Acknowledgement}
We thank the reviewers for their valuable feedback on this work. This research was supported in part by NSF Award (No. 2117439).

\bibliography{causal_rag}

\clearpage
\appendix

\section*{Appendix}

\section{LLM Prompts Used} \label{sec:appendixA}

\begin{figure*}[b]
    \centering
    \includegraphics[width=1\linewidth, trim=25mm 31mm 115mm 13mm, clip]{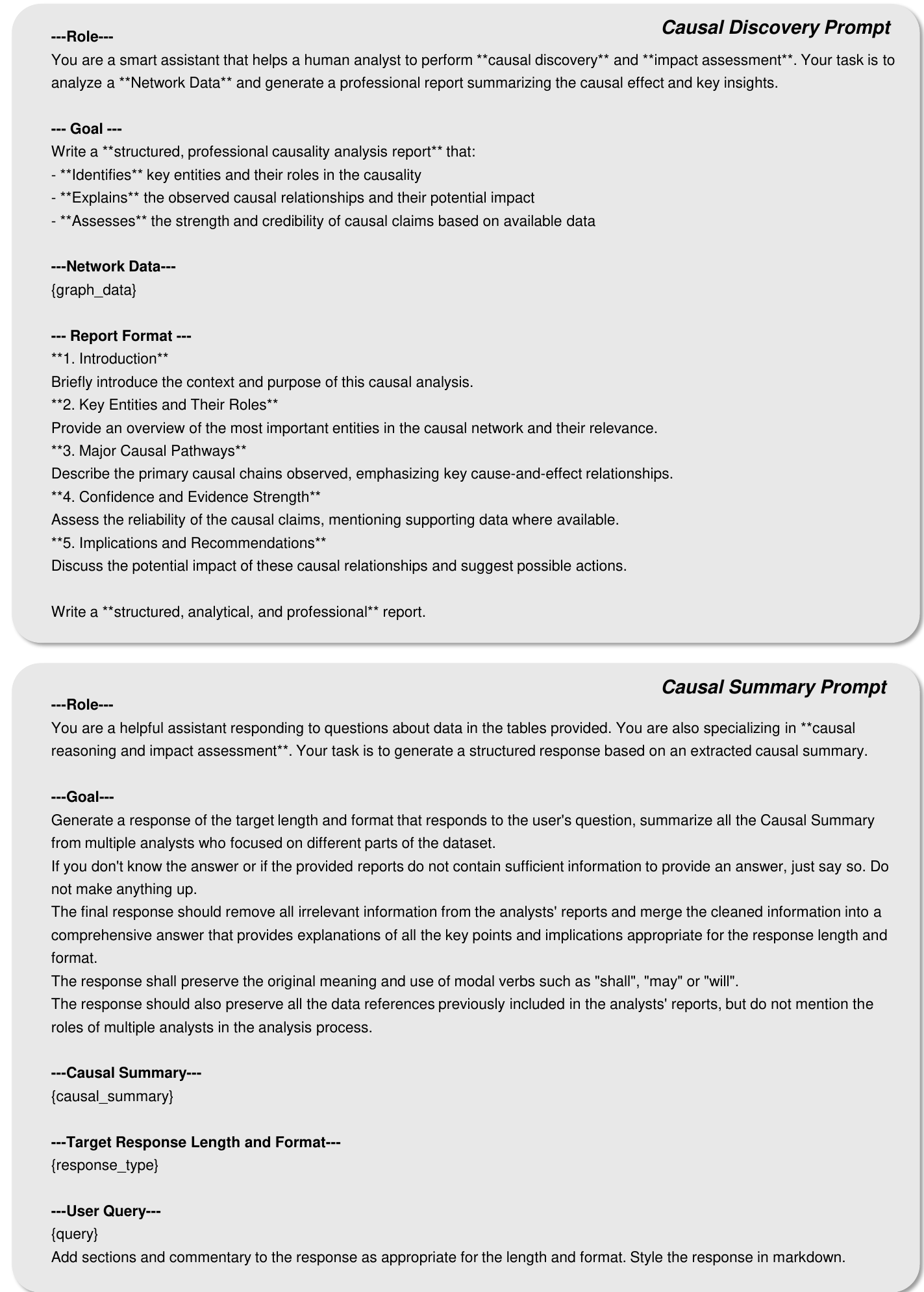}
    \captionsetup{aboveskip=5pt, belowskip=0pt}
    \caption{LLM prompts for causal discovery and causal summary}
    \label{fig7:prompts}
\end{figure*}

For transparency we reproduce the exact templates used in our pipeline (see Figure \ref{fig7:prompts}).

\subsection{Causal–discovery prompt}
\begin{flushleft}\small
\verb|---Role---| You are a smart assistant …\\
\verb|--- Goal ---| Write a structured, professional causality analysis report …\\
\verb|---Network Data---| \{graph\_data\}\\
\verb|--- Report Format ---| … (\textit{see listing in the figure})
\end{flushleft}

\subsection{Causal–summary prompt}
\begin{flushleft}\small
\verb|---Role---| You are a helpful assistant …\\
\verb|---Goal---| Generate a response … merge the cleaned information …\\
\verb|---Causal Summary---| \{causal\_summary\}\\
\verb|---Target Response Length and Format---| \{response\_type\}\\
\verb|---User Query---| \{query\}
\end{flushleft}

The prompts follow three practical heuristics: (i) role
conditioning to steer style; (ii) explicit section headings to
simplify post-processing; (iii) a strict token budget to
keep latency manageable.

\section{OpenAlex Dataset Construction} \label{sec:appendixB}

We sample journal and conference papers with full text from the
OpenAlex API and stratify by top-level field to avoid
domain bias (see Table \ref{tab:domain_token_distribution}).

For each document we ask an LLM to draft \textit{five} answerable questions, inspired by the procedure in \citet{edge_local_2024}. Table~\ref{tab:openalex_case} in this appendix shows one complete example of the resulting \{document, questions\} record.

\begin{table*}[b]
\small
\centering
\renewcommand{\arraystretch}{1.2}
\begin{tabular}{@{}p{3cm} P{11cm}@{}}
\toprule
\textbf{Dataset} & \textbf{OpenAlex~\cite{priem_openalex_2022}} \\ \midrule
Discipline & Atomic Physics \\
Author & Thom H.~Dunning (1989) \\
Text &
In the past, basis sets for correlated molecular calculations were largely taken from single-configuration calculations. Recently, Almlöf, Taylor, and co-workers showed that atomic natural orbitals (ANOs) provide an excellent description of correlation effects. Here we report a careful oxygen-atom study establishing that compact primitive Gaussian functions effectively describe correlation when their exponents are optimized. Tests on oxygen-containing molecules indicate these functions perform as well as the ANO sets of Almlöf and Taylor. Guided by the oxygen results, basis sets were developed for all first-row atoms (B–Ne) and hydrogen. Incremental energy lowerings due to correlating functions fall into distinct groups, leading to the concept of consistent sets. The most accurate set, [5s 4p 3d 2f 1g], consistently yields 99\% of the correlation energy obtained with the next larger set, even though the latter contains 50\% more primitives and twice as many polarization functions. For boron, this set recovers 94–97\% of the total (HF+1+2) correlation energy. \\
Questions &
\begin{enumerate}[label=\arabic*., leftmargin=1.2em, itemsep=2pt]
    \item What recent findings support the use of atomic natural orbitals in molecular calculations?
    \item How do compact primitive Gaussian functions contribute to describing correlation effects in oxygen?
    \item Why is exponent optimization important in Gaussian-based calculations?
    \item How do the new first-row basis sets compare in energy lowering from correlation effects?
    \item What accuracy (\%) is achieved for boron with the most compact set, and how does this relate to the number of polarization functions?
\end{enumerate} \\
\bottomrule
\end{tabular}
\caption{Example document metadata, full-text excerpt, and evaluation questions used in our study.}
\label{tab:openalex_case}
\end{table*}

\section{Implementation Details I – System Configuration} \label{sec:appendixC}

\begin{itemize}[leftmargin=1.2em,itemsep=2pt]
  \item \textbf{Base LLM}. All retrieval-time and evaluation calls use
        \texttt{gpt-4o-mini} via the standard OpenAI API.
  \item \textbf{Embeddings}. Chunks and queries are encoded with sentence-transformers \texttt{all-MiniLM-L6-v2} (384 dimension, cosine similarity).
  \item \textbf{Vector store}. Embeddings are stored in a FAISS index; retrieval depth \(k\!=\!3\) throughout.
  \item \textbf{Orchestration}. Pipeline components are wired together with LangChain v0.2; we use its async OpenAI client to parallelise requests.
\end{itemize}

\section{Implementation Details II – Dataset and Evaluation} \label{sec:appendixD}

\paragraph{Question generation.}
For every document we instruct the LLM to produce five \emph{grounded} questions whose answers appear verbatim in the text.  
Constraining the prompts this way keeps the QA task focussed on retrieval quality rather than open-ended speculation.

\paragraph{Graph creation.}
To assess reasoning on structured knowledge, we constructed individual graphs for each document.  
Although the per-document build is computationally expensive, it largely eliminates information bleed between external knowledge and gives every system the same, self-contained knowledge base.

\paragraph{Evaluation workload.}
The experiment therefore consists of disjoint \{document, question, graph\} triples.  
For each triple we evaluate five variants running one retrieval step, one generation step, the $n$ questions, and three automatic metrics (faithfulness, context-precision, context-recall) implemented with the Ragas framework.  
This uniform process enables a fair, like-for-like comparison of different RAGs' reasoning ability.

\end{document}